\documentclass{article}

    \usepackage[preprint]{neurips_2025}

\usepackage[utf8]{inputenc} 
\usepackage[T1]{fontenc}    
\usepackage{hyperref}       
\usepackage{url}            
\usepackage{booktabs}       
\usepackage{amsfonts}       
\usepackage{graphicx}       
\usepackage{nicefrac}       
\usepackage{microtype}      
\usepackage{xcolor}         
\usepackage{caption}

\usepackage{listings}
\definecolor{codegreen}{rgb}{0,0.6,0}
\definecolor{codegray}{rgb}{0.5,0.5,0.5}
\definecolor{codepurple}{rgb}{0.58,0,0.82}
\definecolor{backcolour}{rgb}{0.95,0.95,0.92}
\lstdefinestyle{mystyle}{
    backgroundcolor=\color{backcolour},   
    commentstyle=\color{codegreen},
    keywordstyle=\color{magenta},
    numberstyle=\tiny\color{codegray},
    stringstyle=\color{codepurple},
    basicstyle=\footnotesize\ttfamily,
    breakatwhitespace=false,         
    breaklines=true,                 
    captionpos=b,                    
    keepspaces=true,                 
    numbers=left,                    
    numbersep=5pt,                  
    showspaces=false,                
    showstringspaces=false,
    showtabs=false,                  
    tabsize=2,
    language=Python
}
\usepackage{algorithm}
\usepackage{algpseudocode} 

\usepackage{multirow}
\usepackage{tabularx}
\usepackage{makecell}
\usepackage{rotating}

\usepackage{amsmath} 

\usepackage{graphicx}   
\usepackage{adjustbox}  
\usepackage{caption}    
\usepackage{wrapfig}  
\usepackage{makecell}

\usepackage{amsmath,amsfonts,bm}

\def\eqref#1{equation~\ref{#1}}

\def\1{\bm{1}}

\DeclareMathAlphabet{\mathsfit}{\encodingdefault}{\sfdefault}{m}{sl}
\SetMathAlphabet{\mathsfit}{bold}{\encodingdefault}{\sfdefault}{bx}{n}

\title{InfLVG: Reinforce Inference-Time Consistent\\ Long Video Generation with GRPO}

\author{%
  Xueji Fang\hspace{2em}
  Liyuan Ma$^{\ddagger}$\hspace{2em}
  Zhiyang Chen\hspace{2em}
  Mingyuan Zhou$^\S$\hspace{2em}
  Guo-jun Qi$^{\ddagger}$ \\[1em]
  MAPLE Lab, Westlake University \\
  \texttt{\{fangxueji, maliyuan, chenzhiyang, zhoumingyuan\}@westlake.edu.cn} \\
  \texttt{guojunq@gmail.com}
}

\begin{document}

\maketitle

\begingroup
\renewcommand{\thefootnote}{\fnsymbol{footnote}}
\footnotetext[4]{Work done at Westlake University.} 
\footnotetext[3]{Corresponding author.}               
\endgroup

\begin{abstract}
Recent advances in text-to-video generation, particularly with autoregressive models, have enabled the synthesis of high-quality videos depicting individual scenes. However, extending these models to generate long, cross-scene videos remains a significant challenge. As the context length grows during autoregressive decoding, computational costs rise sharply, and the model's ability to maintain consistency and adhere to evolving textual prompts deteriorates.
We introduce InfLVG, an inference-time framework that enables coherent long video generation without requiring additional long-form video data. InfLVG leverages a learnable context selection policy, optimized via Group Relative Policy Optimization (GRPO), to dynamically identify and retain the most semantically relevant context throughout the generation process.
Instead of accumulating the entire generation history, the policy ranks and selects the top-$K$ most contextually relevant tokens, allowing the model to maintain a fixed computational budget while preserving content consistency and prompt alignment.
To optimize the policy, we design a hybrid reward function that jointly captures semantic alignment, cross-scene consistency, and artifact reduction.
To benchmark performance, we introduce the Cross-scene Video Benchmark (CsVBench) along with an Event Prompt Set (EPS) that simulates complex multi-scene transitions involving shared subjects and varied actions/backgrounds. Experimental results show that InfLVG can extend video length by up to 9×, achieving strong consistency and semantic fidelity across scenes.
Our code is available at \href{https://github.com/MAPLE-AIGC/InfLVG}{https://github.com/MAPLE-AIGC/InfLVG}.
\end{abstract}

\section{Introduction}

\begin{figure}[htpb]
  \centering
    \includegraphics[width=\linewidth]{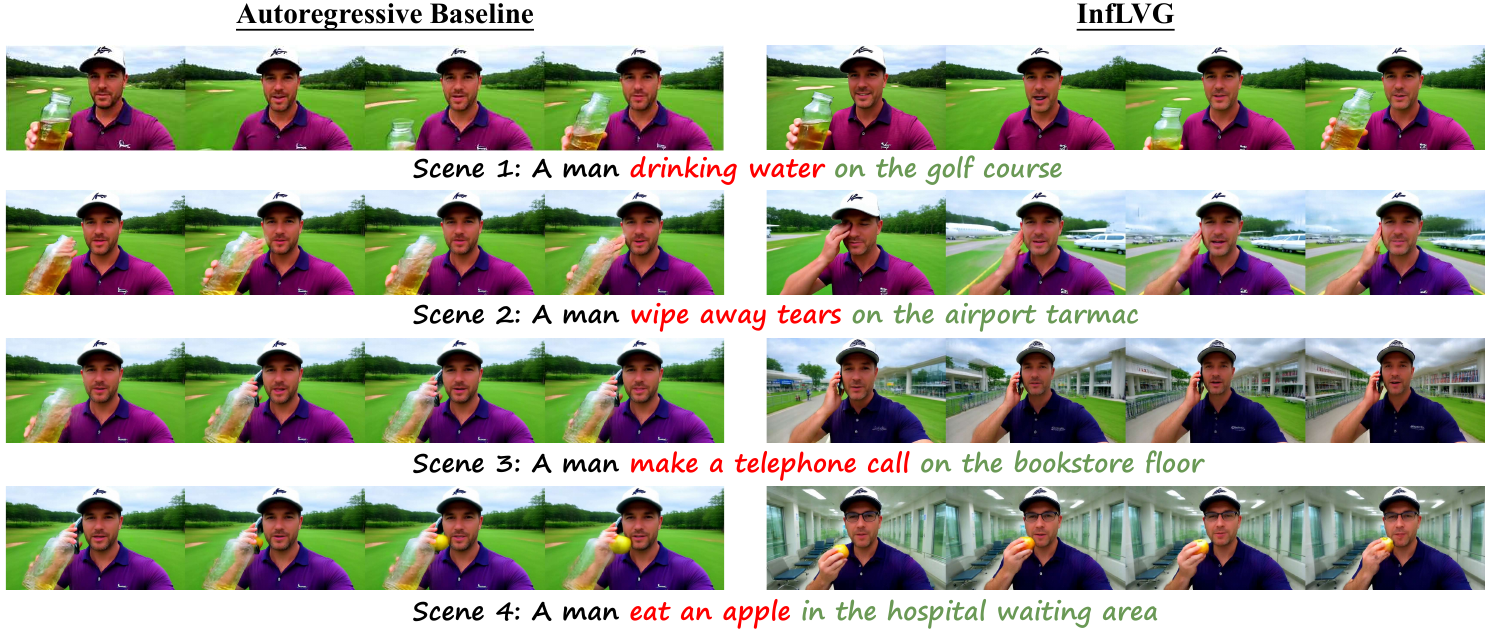}
    \caption{
Challenges in autoregressive long video generation across scenes. (Left) Baseline models tend to repeat initial scene elements (e.g., background, cup) due to unfiltered context accumulation, failing to follow new prompts. (Right) InfLVG addresses this by selectively preserving relevant context, achieving both better prompt alignment and content consistency -- generating both face and environmental elements in accordance with the prompt and given context.
    }
  \label{fig:problems}
\end{figure}

Recent advances in diffusion models~\cite{DDPMs,DDIM,FlowMatching} and transformer-based architectures~\cite{DiT,SD3,DiT-MoE} have significantly advanced video generation. 
While state-of-the-art approaches~\cite{Sora,opensoraplan,CogVideoX,WanX} exhibit impressive visual fidelity and semantic alignment with textual prompts, their extension to longer videos remains constrained by the quadratic computational cost scaling with sequence length and the scarce availability of large-scale, high-quality training data for extended video sequences. 

Most current works based on bidirectional attention mechanisms~\cite{CogVideoX,WanX,Mochi} are trained on fixed-length video segments, inherently limiting their capacity for variable-length extrapolation. Autoregressive video generation~\cite{NOVA,PyFlow,CausVid}, while a seemingly straightforward approach for temporal extension through iterative next-frame prediction, suffers critical limitations in cross-scenario generation. 
As illustrated in Figure \ref{fig:problems}, naively autoregressive extension through frame-by-frame denoising often fail to adapt to new prompt description, as the model remains anchored to initial scene semantics.
The expanding context window accumulates increasingly irrelevant features that dominate the model's attention, effectively suppressing its ability to focus on the specific context required for coherent scene generation.
This necessitates an adaptive selection mechanism that selectively preserves content relevant to scene transitions while filtering irrelevant elements, thereby achieving an optimal balance between content continuity and prompt adherence.

\begin{figure}[htpb]
  \centering
\includegraphics[width=\linewidth]{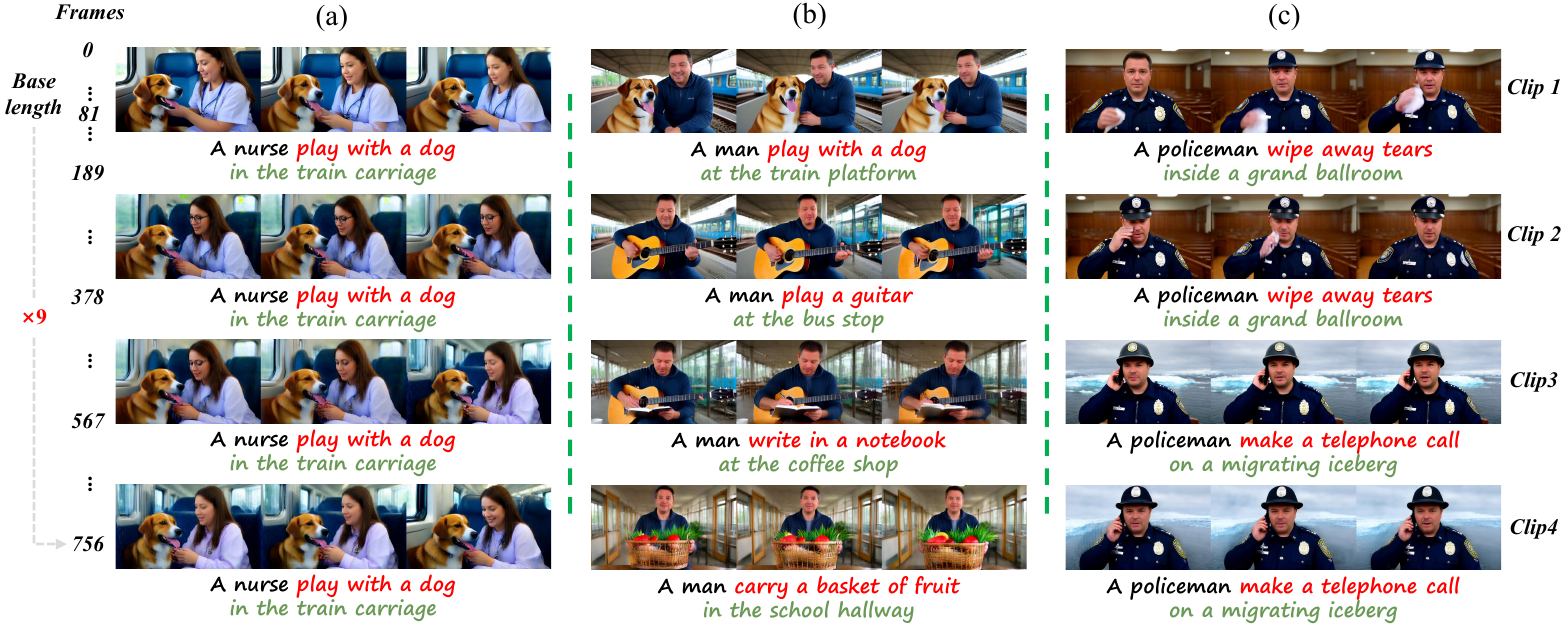}
  \caption{Different scene extension paradigms with InfLVG.
  (a) Single-scene extension, (b) Multi-scene transition with contextual awareness, and (c) Both single- and multi-scenes.
  }
  \label{fig:demo_single_multi}
\end{figure}

To address these issues, we propose \textbf{InfLVG}, an inference-time framework that adaptively adjusts the video context for each individual sample without requiring additional long-form training data.
InfLVG introduces a context selection policy trained via GRPO, guided by hybrid rewards specifically designed to ensure content consistency and semantic alignment with the prompt.
During autoregressive video generation, the policy assigns relevance scores to each context token, estimating their semantic contribution to the newly generated scene relative to preceding content.
Based on these scores, the model performs top-$K$ ranking to select the most relevant context features and uses them to progressively denoise subsequent frame sets.
The reward function is carefully designed to jointly capture content consistency by preserving scene identity across transitions, ensure prompt alignment through accurate text-video semantic matching, and suppress artifacts by penalizing visual distortions in the generated frames.
Through GRPO-based optimization, the policy is reinforced to prioritize context tokens that are critical for cross-scene coherence, while effectively filtering out irrelevant elements.
Moreover, by enforcing a fixed-length context window during inference, our method maintains a bounded computational cost, even when generating long, multi-scene videos.
As illustrated in Figure~\ref{fig:demo_single_multi}, InfLVG enables flexible generation paradigms, including both single-scene extension and multi-scene transitions with contextual awareness.
To evaluate cross-scene video generation performance, we propose CsVBench, a benchmark that constructs multi-scene descriptions featuring the same subject identity across varied actions and backgrounds.

Our contributions are threefold.
First, we propose an inference-time framework for generating multi-scene coherent long videos that effectively balances cross-scene consistency and alignment with dynamic textual prompts without any additional training on long-form video data.
To support this, we design a GRPO-optimized context selection policy guided by hybrid rewards incorporating multi-faceted quality signals. This policy enables the model to identify and emphasize semantically relevant context tokens, thereby maintaining consistency between newly described scenes and previously generated content.
Furthermore, by applying top-$K$ ranking to select a fixed number of context features, our approach maintains a constant computational footprint while allowing the autoregressive model to scale to arbitrarily long video sequences. Experimental results show that our method achieves up to a $9 \times$ increase in video length while preserving high generation quality.

\section{Related Work}
\textbf{Video Generation Models.} 
Recent years have witnessed substantial advances in video generation, driven by the development of diffusion and autoregressive generative modeling.
A seminal breakthrough emerged through the scaling laws in the pretraining of Diffusion Transformer (DiT) architecture~\cite{DiT, DiT-MoE}, where video generation quality exhibits remarkable improvements with increased model parameters and training data scale.
However, conventional pretraining paradigms that condition on single text-video pairs~\cite{Sora, CogVideoX, WanX, Mochi} inherently limit these models to generating videos within a single scene or narrative context.
Parallel efforts to extend generation length without retraining attempt employ sliding window mechanism~\cite{FreeNoise, FIFO, Ouroboros}, enabling longer video sequences through local temporal context conditioning. 
However, these methods inherently depend on the quality of the pretrained backbone and often produce spatiotemporal inconsistencies in DiT-based frameworks, particularly when synthesizing dynamic scene transitions. 
Alternative approaches~\cite{NOVA, PyFlow, CausVid} integrate autoregressive modeling with diffusion processes, enabling theoretically unbounded generation lengths through iterative frame prediction. Nevertheless, their training protocols are typically optimized for single-scene generation and thus struggle to generalize to cross-scene narratives, while expanding training data to diverse scenarios incurs prohibitive computational costs.
Our framework proposes a reinforcement learning framework built upon autoregressive video diffusion models, achieving consistent multi-scene generation and faithful adherence to new textual prompts without requiring additional training data. This is realized through adaptive history context selection and hybrid reward optimization.

\textbf{Aligning Visual Generation with Preference Feedback.} 
Building upon the success of Reinforcement Learning from Human Feedback (RLHF)~\cite{DeepSeek-r1,CoT-Survey-Hao-Fei,JudgeLRM,Sketch-of-Thought,s1} in aligning language models with human preferences has inspired analogous approaches for diffusion-based generation.
Pioneering works~\cite{DDPO, DPOK, Diffusion-DPO} established the foundation by employing policy gradient methods and direct preference optimization~\cite{DPO} to align text-to-image diffusion models with perceptual quality metrics like Imagereward~\cite{ImageReward}, HPSV2~\cite{hps, HPSv2}, and PickScore~\cite{pickscore}.
Building upon these foundations, subsequent research has extended this paradigm to video generation.
InstructVideo~\cite{InstructVideo} first adapted image reward models through RL-driven fine-tuning of video diffusion models. 
Further studies~\cite{wu2024boosting, VideoReward, VideoPrefer, GRADEO, VideoScore} have designed video-specific reward functions and leveraged reinforcement learning to fine-tune text-to-video models, thereby significantly enhancing the dynamic quality of generated videos.

\section{Method}
To address the challenge of content inconsistency in autoregressive video generation, we present the designs of the policy and reward models within the reinforcement learning framework at inference time.
Section ~\ref{sec:pre} reviews the preliminaries of autoregressive video diffusion and flow matching models.
Section ~\ref{sec:kvsel} presents the architecture of context selection model and top-$K$ ranking for KV cache sampling during GRPO training. Section ~\ref{sec:hybrid} details the hybrid reward functions for jointly optimizing video quality, textual alignment, and visual consistency.

\begin{figure}[htpb]
  \centering
    \includegraphics[width=\linewidth]{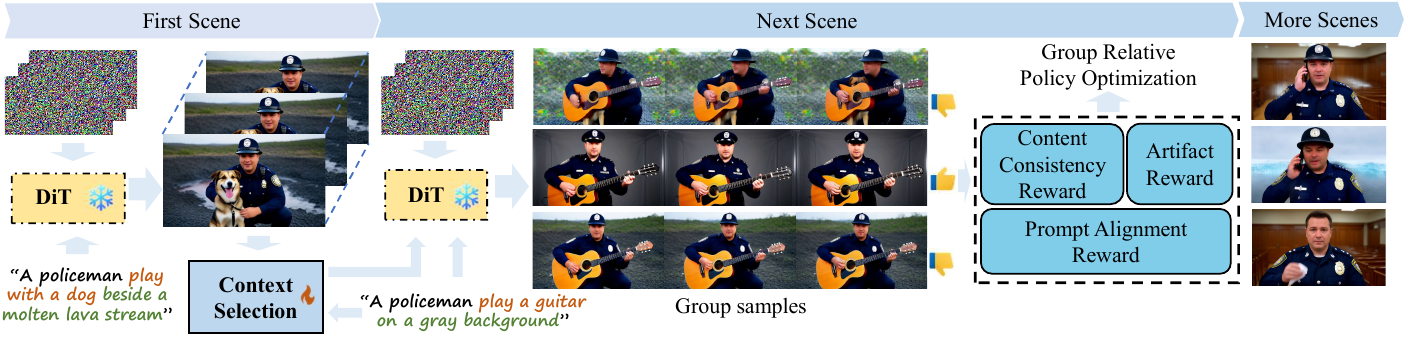}
    \caption{GRPO training pipeline. The DiT-based autoregressive video model generates a group of next scenes under top‐$K$ sampling actions. These videos are scored by the hybrid rewards and InfLVG utilizes GRPO to update the context selection model.}
  \label{fig:grpo}
\end{figure}

\subsection{Preliminaries}
\label{sec:pre}
\textbf{Autoregressive Text-to-Video Generation.}
Autoregressive text-to-video diffusion models have demonstrated significant potential for scalable long-duration video synthesis through temporal autoregressive modeling in latent space.
These architectures employ a next-frame set prediction framework where temporal causality enforces a causal attention mechanism -- the $l^\text{th}$ video segment $V_{nl:n(l+1)-1}$ with $n$ frames conditions exclusively on preceding content $V_{<nl}$.
Formally, the joint video distribution given a text prompt $P$ decomposes below,
\begin{equation}
    p(V_{0:n-1}, ...,V_{n(N-1):nN-1}) = \prod_{l=0}^{N-1}p(V_{nl:n(l+1)-1} | \mathcal{F}_{\theta}(V_{<nl}), P),
\label{eq:arv}
\end{equation}
where $N$ video segments are generated and the frame set size can be a single frame~\cite{PyFlow, NOVA} or multiple frames~\cite{CausVid}.
Diffusion Transformer (DiT)~\cite{DiT} is widely adopted in autoregressive video generation, and $\mathcal{F}_{\theta}$ represents the context frames management strategy with learnable parameters $\theta$.

Recent work such as CausVid~\cite{CausVid}, a distilled autoregressive video diffusion model derived from WanX~\cite{WanX}, enables segment-by-segment video generation in just a few denoising steps.
Specifically, 
the few-step generator $\mathcal{G}$ first initializes the $l^\text{th}$ video segment $V^{t_T}_{l} \sim \mathcal{N}(0, I)$ and then iteratively refines it through $T$ denoising steps. This reverse process follows the update rule:
\begin{equation}
V^{t_{j-1}}_{l} = \alpha_{t_{j-1}} \mathcal{G}\big(V^{t_j}_{l}, t_j; \mathcal{F}_{\theta}(V_{<nl}), P)\big) + \sigma_{t_{j-1}} \epsilon, \quad \epsilon \sim \mathcal{N}(0,I),\ \ j=T,\dots,1
\end{equation}
where $V^{t_{j}}_{l}$ denotes the $l^\text{th}$ video segment at timestep $t_j$, $\alpha_{t}$ and $\sigma_{t}$ are predefined noise scheduling coefficients from the distillation process, and $\mathcal{F}_{\theta}(V_{<nl})$ represents the KV cache aggregated from previously generated video frames.
The causal attention in $\mathcal{G}$ ensures temporal consistency by constraining each video segment generation to depend only on preceding frames.

\textbf{GRPO for Autoregressive Video Extension.}
Inspired by the success of reinforcement learning in the post-training of Large Language Models (LLM)~\cite{grpo,visionr1} and generative diffusion models~\cite{DDPO,DPOK,Diffusion-DPO} to align the pretrained base model with human preferences, we utilize the GRPO to reinforce the consistency of the autoregressive video generation with specific designed rewards.
GRPO optimizes the policy $\pi_{\theta }$ by maximizing the advantages of generated video samples.
Given a group of $G$ video $\{V\}_{i=1}^G$ generated through autoregressive frame prediction with iterative latent denoising, GRPO estimates the reward $r_i$ of generated video by measuring the denoised clean frames with specific reward functions, and then calculates the advantages by comparing each other inside the group:
\begin{equation}
    \label{eq:adv}
    A_i = \frac{{r}_i - \text{mean}(\{r_i\}_{i=1}^G)}{\text{std}(\{r_i\}_{i=1}^G)}.
\end{equation}
The optimization objective is defined as follows:
\begin{equation}
    \mathcal{J}_\text{GRPO}(\theta) = \mathbb{E}_{V_i \sim \pi_{\theta_\text{old}}} \left[
    \begin{aligned}
        \frac{1}{G}\sum_{i=1}^{G} \mathrm{min} \left( \frac{\pi_{\theta}(V_i)}{\pi_{\theta_\text{old}}(V_i)} A_i, \mathrm{clip} \left( \frac{\pi_{\theta}(V_i)}{\pi_{\theta_\text{old}}(V_i)}, 1 - \delta, 1 + \delta \right) A_i \right)
    \end{aligned}
    \right],
\label{eq:grpo}
\end{equation}
where $A_i$ denotes the group-wise advantage, $\delta$ denote clipping hyper-parameter~\cite{ppo}.
The rewards are designed by specific problem setting and will be elaborated in following sections.
We remove the KL penalty term from vanilla GRPO objective to eliminate reference model overhead.

\subsection{Inference-time Context Selection for Cross-scene Video Extension}
\label{sec:kvsel}
Current autoregressive video generators struggle with content consistency over long sequences and prompt adherence. To address this, we introduce a dynamic context selection mechanism that masks irrelevant and preserves relevant video features at inference time. As illustrated in Figure~\ref{fig:grpo}, our approach ensures coherent, cross-scene video generation under evolving prompts.
Starting from an initial $n$-frame video segment $V_{0:n-1} \in \mathbb{R}^{nhw}$ generated from prompt $P_1$, the model extends it to a subsequent segment $V_{n:2n-1}$ conditioned on a new prompt $P_2$.
The context selection model is optimized by GRPO with hybrid rewards, which reinforce the probability of video samples that has better identity consistency, text alignment, and less visual artifact.
Subsequently, the framework propagates the generation process to synthesize longer sequences $V_{nl:n(l+1)-1}$ ($l>1$) with more scenes by recursively leveraging the selected context from previously generated video segments, enabling scalable cross-scene construction.
We parameterize the context selection policy using a learnable function $\mathcal{F}_{\theta}$,  which dynamically determines the most relevant tokens to attend to during the video denoising process.  This policy is designed to strike a balance between preserving content continuity and adhering to newly introduced prompt semantics.

\begin{algorithm}[!t] 
  \small
  \caption{Context Tokens Selection via Plackett--Luce~\cite{ragain2018choosing,PlackettLuce} Sampling }
  \label{alg:pl_sampling}
  \begin{algorithmic}[1]
  \Require $L$ video context tokens $V_{<nl}$ with $nl$ frames, prompt of current video segment $P_{l}$, context selection model $\mathcal{F}_\theta$, number of tokens to select $K \ll L$.
  \State \textbf{Compute} score for each token: 
  $s^i_L = \mathcal{F}^i_{\theta}(V_{<nl}, P_{l}) \quad \text{for } i \in \{0,  \dots, L-1\}$
  \State \textbf{Initialize} empty ranking indices list for selected context: $\mathcal{C}_L = [\ ]$
  \State \textbf{Initialize} candidates pool: $\mathcal{U}_L \gets \{0,\dots,L-1\}$
  \For{$k = 0$ to $K-1$}
      \State \textbf{Normalize scores} over remaining context candidates: 
      $
        p^j_k = \left\{
            \begin{array}{ll}
                \displaystyle \frac{\exp(s_{L}^{j})}{\sum_{j' \in \mathcal{U}_L} \exp(s_{L}^{j'})}, & \forall j \in \mathcal{U}_L \\[10pt]
                \vcenter{\hbox{\rule{0pt}{1.2em}}}0, & \forall j \in \mathcal{C}_L
            \end{array}
        \right.
      $
      \State \textbf{Sample} a token $c_L(k) \sim \text{Multinomial}(p_k)$
      \State \textbf{Update selection}: $\mathcal{C}_L \gets \mathcal{C}_L \cup \{c_L(k)\}$
      \State \textbf{Remove candidates} $\mathcal{U}_L \gets \mathcal{U}_L \setminus \{c_L(k)\}$
  \EndFor
  \State \Return top-$K$ context token indices list $\mathcal{C}_L$, the set of unselected tokens $\mathcal{U}_L$
  \end{algorithmic}
\end{algorithm}

\begin{figure}[!t]
  \centering
    \includegraphics[width=\linewidth]{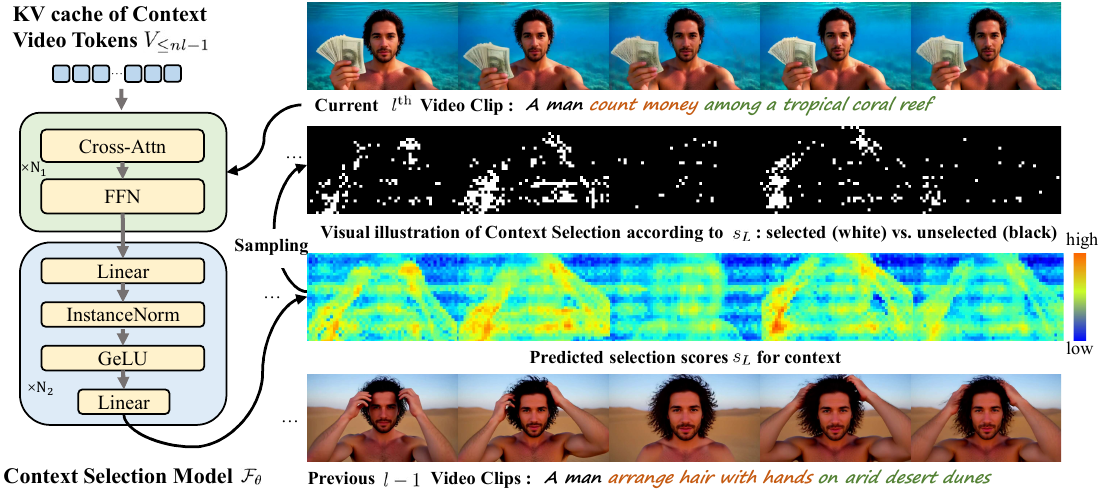}
    \caption{Illustration of Context Selection Model $\mathcal{F}_{\theta}$. Past video tokens and the current prompt are fused via cross-attention, then top-$K$ ranking is applied to sample context from the KV cache.}
  \label{fig:policy}
\end{figure}

\textbf{Context Selection Policy.} 
During the generation of the $l^{\text{th}}$ video segment $V_{nl:n(l+1)-1}$, i.e. the $n$ frames from the $(nl)^{\text{th}}$ to the $(n(l+1)-1)^{\text{th}}$. There already exists $L=nlhw$ history tokens from the preceding segments $V_{<nl}$ containing $nl$ frames, which is too much to be contained in the context. 
Thus, the context selection policy adaptively retrieve and aggregate the most relevant $K$ video context tokens from all these candidates. 
These tokens then provide additional key \& value features for self-attention in the generation of the current video segment.
Specifically, given the textual prompt $P_{l}$ describing the current video generation result and context video features from previous generated video segments $V_{\leq {nl-1}}$, the context selection model $\mathcal{F}_{\theta}$ first predict a score value $s_{L}^i = \mathcal{F}^i_{\theta}\bigl(V_{ \leq {nl-1}},\,P_{l}\bigr)$ for each token $i$ in $L$ context tokens at the current generation step. 
Based on these scores, we can sample a top-$K$ ranking indices list $\mathcal{C}_L=[c_L(1),..., c_L(K)]$ from a Plackett–Luce model~\cite{ragain2018choosing,PlackettLuce} sequentially, and all candidate tokens within $\mathcal{C}_L$ are put into the context. 
The top-$K$ ranking is practically implemented with sampling from a multinomial distribution determined by $s_{L}$ without replacement, such that context tokens with high scores would be picked with high probabilities.
The detailed sampling process is shown in Algorithm~\ref{alg:pl_sampling}.
Formally, let $\mathcal{U}_L$ denote the set of unselected tokens, i.e., it is the complement set $\mathcal{C}_{L}^-$ of $\mathcal{C}_L$.
Then, the probability of sampling a specific ranking list $\mathcal{C}_L$ on the whole context video tokens is given by:
\begin{equation}
    \pi_{\theta}(\mathcal{C}_L|V_{<nl}) = \prod_{k=1}^K
    \frac{
        \exp(s_{L}^{c_L(k)})
    }{
        \sum_{j=k}^{K}\exp(s_{L}^{c_L(j)}) + \sum_{j'\in \mathcal{U}_L}^{}\exp(s_{L}^{c_L(j')})
    },
\end{equation}
This constructs a probabilistic policy that allows to train the model with the GRPO and the other PPO reinforcement learning approaches. 
After determining the ranking indices list $\mathcal{C}_{L}$, we can retrieve the sparse key-value pairs $(\mathbf{K}_{\mathcal{C}_L}, \mathbf{V}_{\mathcal{C}_L})$ from the full context accordingly.
For autoregressive generation of subsequent segment ${V}_{nl:n(l+1)-1}$, attention computation in autoregressive video diffusion transformer operates exclusively on these selected tokens:
\begin{equation}
\text{Attention}(\mathbf{Q}, \mathbf{K}_{\mathcal{C}_L}, \mathbf{V}_{\mathcal{C}_L}) = \text{Softmax}\left(\frac{\mathbf{Q}\mathbf{K}_{\mathcal{C}_L}^\top}{\sqrt{d}}\right)\mathbf{V}_{\mathcal{C}_L},
\end{equation}
where queries $\mathbf{Q}$ originate from current denoising video tokens, $d$ is the dimension of hidden states for each attention head.
By enforcing \( K \ll L\), the attention calculation cost of video segment is bounded, enabling affordable long video generation without quadratic blowup.

\textbf{Context Selection Model Architecture.} 
As illustrated in Figure~\ref{fig:policy}, our selection model processes the key-value cache from preceding video tokens $V_{<nl}$ through $\text{N}_1$ cross-attention blocks that incorporate prompt information from the current video segment to be generated. 
This architecture enables semantic-aware context selection by conditioning on both historical context and future generation semantics. 
The processed tokens then pass through $\text{N}_2$ lightweight linear projection layers and  produce per-token selection scores.

\subsection{Joint Optimization with Hybrid Rewards} 
\label{sec:hybrid}
The core challenge of cross-scene video extension lies in maintaining both content consistency and prompt alignment during autoregressive generation. 
To evaluate the cross-scene content consistency during video extension, we utilize the pretrained face recognition model~\cite{Arc2Face} as the content consistency reward model $r_\text{content}$, which assesses identity preservation by measuring feature similarity between facial embeddings extracted from different scenes. 
Specifically, given the newly generated video segment $V^\text{cur} = V_{nl:n(l+1)-1}$ and the preceding frames $V^\text{prev} =V_{<nl}$, we uniformly sample $E$ keyframes from both segments and compute average pairwise cosine similarity as:
\begin{equation}
r_{\text{content}}(V_{nl:n(l+1)-1}, V_{<nl}) = \frac{1}{E^2} \sum_{i=1}^{E} \sum_{j=1}^{E} \text{sim}\big( \phi({V}^{\text{cur}}_j), \phi({V}^{\text{prev}}_i)\big),
\end{equation}
where $\phi(\cdot)$ denotes the ArcFace feature extractor, and $\text{sim}(\cdot,\cdot)$ represents cosine similarity.
To prevent degenerate solutions where the context selection policy trivially retains all tokens and repeat the previous scene while ignoring the new prompt, we introduce a prompt-alignment reward $r_\text{clip}$ to encourage fidelity to the current prompt.
This is computed as the average CLIP similarity between $Q$ uniformly sampled frames from $V_{nl:n(l+1)-1}$ and the new prompt $C_{l+1}$:
\begin{equation}
r_{\text{clip}}(C_{l+1}, V_{nl:n(l+1)-1}) = \frac{1}{Q} \sum_{i=1}^{T} \text{sim}\big(\psi(C_{l+1}), \psi(V_i^\text{cur})\big),
\end{equation}
where $\psi(\cdot)$ is the CLIP embedding function, and ${V}^\text{cur}_i$ denotes the sampled frame.
We observe that conditioning on poorly selected context can cause the model to produce blocky, mosaic-like color artifacts that degrade visual quality (see details in the Appendix). To mitigate this, we leverage a Vision-Language Model (VLM) to detect artifacts through binary classification using a specially designed instruction prompt. The artifact reward $r_\text{artifact}$ is set to 1 if no artifact is detected, and 0 otherwise.
The final hybrid reward is a summation of all three components:
\begin{equation}
    r = r_{\text{content}} + r_{\text{clip}} + r_{\text{artifact}}.
\end{equation}
Combining the hybrid rewards and the context selection policy, we can update the context selection model $\mathcal{F}_{\theta}$ by maximizing the objective defined in Equation~\ref{eq:grpo}.

\begin{table}[th]
 \centering
\caption{Quantitative comparison of various context selection mechanism designs and our proposed InfLVG. Overall consistency score is calculated as the average of the Text-Video Alignment and Cross-Scene Consistency scores. The best, second best, and third best results are highlighted in \textbf{bold}, \underline{underlined}, and \textit{italicized} fonts, respectively. }
 \label{tab:compare}
 \resizebox{\textwidth}{!}{%
 \begin{tabular}{@{}l*{10}{c}@{}}
 \toprule
 \multirow{2}{*}{\textbf{{Model Variants}}} &
 \multicolumn{3}{c}{\textbf{Video Quality $\uparrow$}} &
 \multicolumn{3}{c}{\textbf{Text–Video Alignment $\uparrow$}} &
 \multicolumn{3}{c}{\textbf{Cross–Scene Consistency $\uparrow$}} &
 \multirow{2}{*}{\makecell{\textbf{Overall consistency} \\ \textbf{score} $\uparrow$}} \\
 \cmidrule(lr){2-4} \cmidrule(lr){5-7} \cmidrule(lr){8-10}
 & HPSv2 & Aesthetic & QWen & CLIP–Flan & ViCLIP & QWen & ArcFace-42M & ArcFace-360K & QWen & \\
 \midrule
 Vanilla         & 0.2380            & \textbf{4.9910}    & 0.9751             & 0.3180             & 0.1176         & 0.2007             & \textbf{0.2468}    & \textbf{0.2578}    & \textbf{0.9971}    & 0.3563 \\
 Rand. Per-token & 0.2680            & 4.7700             & 0.9776             & {0.5427} & 0.1998         & \textit{0.5500}    & 0.0947             & 0.1065             & \underline{0.9471} & 0.4068 \\
 Rand. Per-frame & 0.2680            & 4.7260             & 0.9760             & 0.5364    & 0.1984         & 0.5169             & 0.1279             & 0.1413             & 0.9265             & 0.4079 \\
 Sliding Window  & \textit{0.2710}   & 4.5250             & 0.8120             & \textbf{0.6350}    & \textbf{0.2261} & \textbf{0.7434}   & 0.0701             & 0.0750             & 0.6853             & 0.4058 \\
 Global-local    & \textbf{0.2730}   & \underline{4.9150} & \underline{0.9947} & 0.5297             & 0.1989         & 0.5206             & \textit{0.1431}    & \textit{0.1582}    & 0.9088             & \textit{0.4099} \\
 Rand. Global-local & \textbf{0.2730} & \textit{4.8500}    & \textit{0.9838}    & \underline{0.5538}             & \textit{0.1999} & \underline{0.5588} & 0.1204             & 0.1337             & \underline{0.9471} & \underline{0.4190} \\
 \midrule
 Ours            & \underline{0.2720} & 4.6990             & \textbf{0.9983}    & \textit{0.5523 }            & \underline{0.2003} & 0.5484         & \underline{0.1684} & \underline{0.1825} & \textit{0.9281}    & \textbf{{0.4300}} \\
 \bottomrule
 \end{tabular}%
 }
\end{table}

\begin{figure}[!th] 
    \centering
    \includegraphics[width=\linewidth]{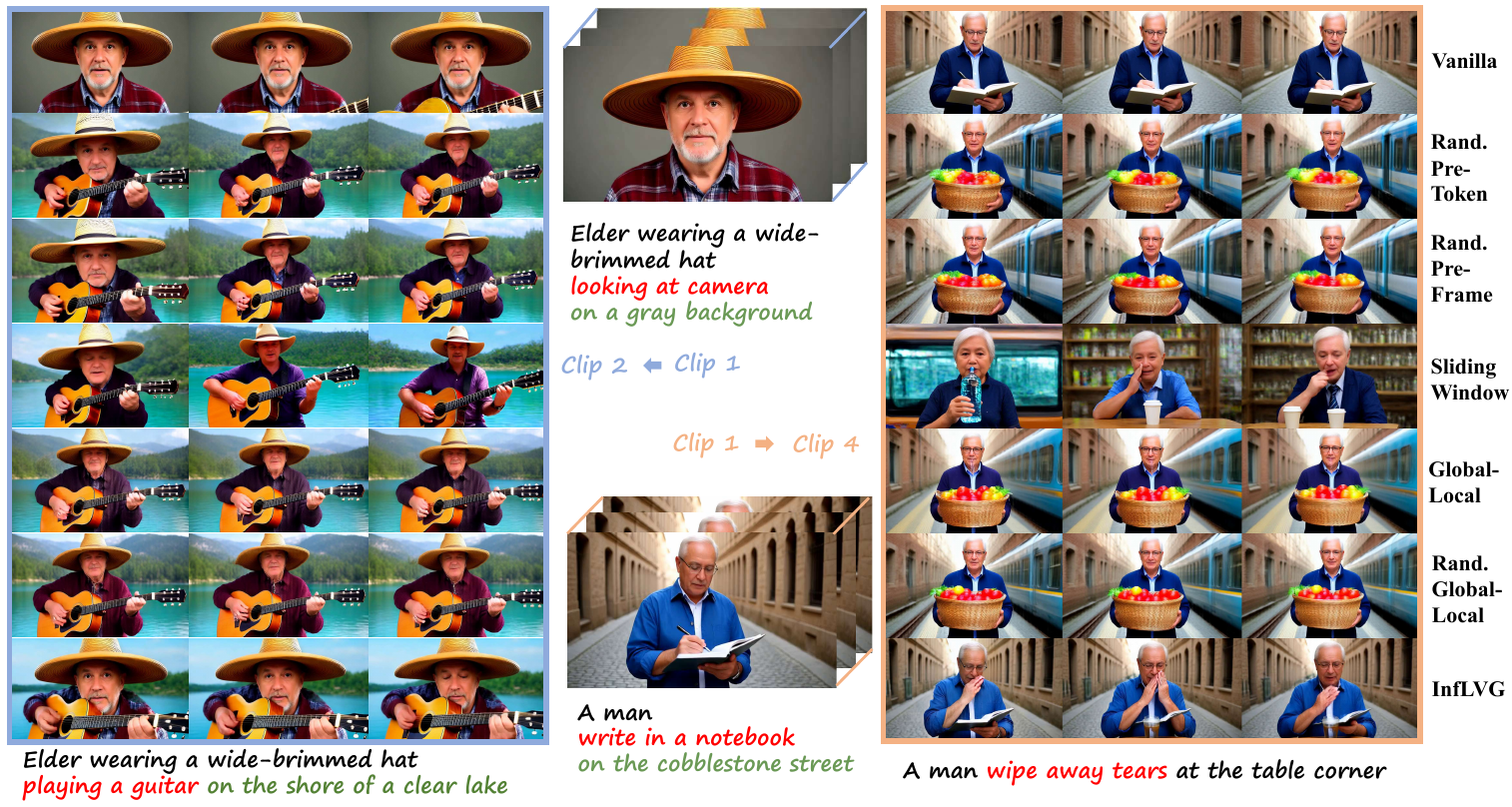}
    \caption{Comparison of different context selection designs under cross-scene video generation .
    }
    \label{fig:comp}
\end{figure}

\section{Experiment}
\label{sec:exp}

\subsection{Implementation Setup}
\textbf{Inference-time GRPO Training Details.}
Our experiments are built upon CausVid\cite{CausVid}, which is capable of generating 5-second videos (81 frames at 16 fps) using $T=3$ denoising steps. In our experiments, we extended the video length by $ 9 \times$, resulting in 765 frames across 4 scenes. The content consistency and prompt alignment rewards were calculated with $E=8$ and $Q=16$. During GRPO training, we set the group size to 10 to ensure thorough exploration. We employed the AdamW optimizer with a constant learning rate of 0.001. For each scene generation, we optimized the policy model for 20 iterations during inference. 
In the context selection model, we set $\text{N}_1 = 1$ and $\text{N}_2 = 2$. All experiments are conducted on NVIDIA H800 GPUs.

\textbf{Cross-scene Video Benchmark}.
To evaluate the text adherent, cross clip consistency, and video quality, we proposed CsVBench, a benchmark designed for cross-scene video generation.
The evaluation prompts are organized as Event Prompts Sets in the structured format "[Human] [Action] [Background]". 
We use an LLM\cite{Gemini} to generate a pool of common concepts, comprising 12 human identities, 16 actions, and 90 backgrounds. 
Cross-scene prompts are generated by randomly pairing different actions and backgrounds while keeping the human identity fixed, enabling assessment of identity consistency across scenes.
The final benchmark was curated  by generating and filtering candidate EPS triplets, with details provided in the Appendix. All experiments were conducted on this filtered set. 

To quantitatively evaluate our method, we assess Video Quality, Text-Video Alignment and Cross-Scene Consistency using a suite of metrics. These include HPSv2~\cite{HPSv2}, Aesthetic Score~\cite{Laion-5B-AesScore}, CLIP-Flan~\cite{CLIPFlan}, ViCLIP~\cite{InternVid}, and two ArcFace variants (ArcFace-42M~\cite{Arc2Face}, and ArcFace-360K~\cite{ArcFace}) for face identity similarity.
We also incorporate QWen2.5-VL~\cite{QWen2.5} for evaluation across these dimension. Full evaluation details and metric definitions are provided in Appendix.

\subsection{Performance Analysis}
\begin{figure}[!th]
    \centering
    \includegraphics[width=\linewidth]{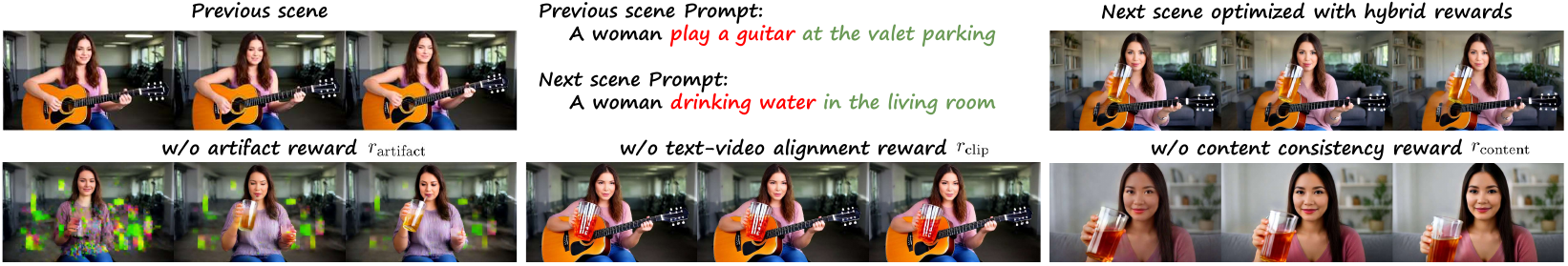}
    \caption{Ablation study on hybrid rewards.}
    \label{fig:ab_reward}
\end{figure}
We compare our proposed InfLVG against various context selection strategies for data-free cross-scene autoregressive video generation.
Initially, we employ the original autoregressive model for long video generation beyond its training length by naively attending to all historical context video tokens based on the text description; we refer to this as the \textbf{Vanilla Extension}.
To investigate the significance of accurate and relevant context selection for consistent cross-scene video generation, we design the \textbf{Random Per-token} and \textbf{Random Per-frame} context selection mechanisms, which construct historical video context by randomly selecting tokens and frames, respectively. 
Furthermore, we implement a \textbf{Sliding Window} selection method, allowing the model to attend to a fixed number of locally adjacent frames during video extension, mirroring similar approaches in prior work (e.g., FreeNoise~\cite{FreeNoise}).
Inspired by the sparse attention in BigBird~\cite{BigBird}, we design two hybrid context selection methods that combine both global and local context: \textbf{Global-Local } selection enforces attention to the first frame as a global anchor and a set of temporally adjacent local frames from recent segments; \textbf{Random Global-Local} selection further introduces stochastic sampling of intermediate frames from earlier segments. Notably, all selection strategies, including our InfLVG, are configured to select the same number of context video tokens, $6hw$, ensuring a fair comparison.

As illustrated in Figure~\ref{fig:comp}, Vanilla Extension results in text misalignment while paradoxically maintaining a similar semantic visual to the first clip. 
With the generation of more clips, the initial prompt's visual features persist, leading to high face consistency but low text alignment scores as presented in Table~\ref{tab:compare}. 
Randomly selecting context in token- or frame-wise yields nearly identical visual outcomes with minimal quantitative differences, yet both fail to generate well text-aligned videos, especially with increasing clip generation, as evidenced by the persistent fruit holding in the right panels. 
Selection with sliding window mechanism focuses solely on local context exhibits poor content consistency due to its inability to capture long-range dependencies. 
Global-Local and Random Global-Local also struggle to effectively balance text alignment and human consistency, as they cannot follow the new prompt describing the new action of human. 
In contrast, our proposed InfLVG demonstrates superior flexibility in dynamically selecting semantically relevant context. It effectively maintains both visual consistency and prompt adherence, achieving the best overall consistency score in Table~\ref{tab:compare}. Qualitatively, InfLVG preserves human identity across scenes while accurately following scene transitions prompted in the text as presented in Figure~\ref{fig:comp}.

\textbf{Effect of different rewards.}  
As shown in Figure~\ref{fig:ab_reward} , removing the artifact suppression reward $r_\text{artifact}$ leads the policy to select irrelevant background context, resulting in noticeable visual artifacts.
When the text-video alignment reward $r_\text{clip}$ is omitted, the generated video fails to fully comply with the prompt.
For example, the "drinking water" action occurs, but not in the correct "living room" setting.
Additionally, excluding the content consistency reward $r_\text{content}$ disrupts the temporal continuity of the depicted subject, causing inconsistencies in appearance and identity across scenes.

\begin{wraptable}[14]{c}{0.4\textwidth}
  \centering
  \resizebox{\linewidth}{!}{%
    \begin{tabular}{@{}lccc@{}}
      \toprule
      \textbf{Context Length} & $K{=}3hw$ & $K{=}6hw$ & $K{=}12hw$ \\
      \midrule
      \multicolumn{4}{@{}l}{\textbf{Text–Video Alignment} $\uparrow$} \\
        \cmidrule(lr){1-4}
        CLIP–Flan     & 60.51 & 56.47 & 50.14 \\
        ViCLIP        & 20.08 & 19.71 & 18.26 \\
        QWen          & 66.62 & 57.21 & 47.06 \\
        \midrule
        \multicolumn{4}{@{}l}{\textbf{Cross–Scene Consistency} $\uparrow$} \\
        \cmidrule(lr){1-4}
        ArcFace–42M   & 6.93  & 12.19 & 15.12 \\
        ArcFace–360K  & 7.80  & 13.04 & 16.37 \\
        QWen          & 80.59 & 86.76 & 93.82 \\
        \bottomrule
    \end{tabular}%
  }
  \caption{Numerical comparison between models with different context length in sampling with top-$K$ ranking.}
  \label{tab:ab_context}
\end{wraptable}

\begin{figure}[htpb]
    \centering
    \vspace{-10pt}
    \includegraphics[width=\linewidth]{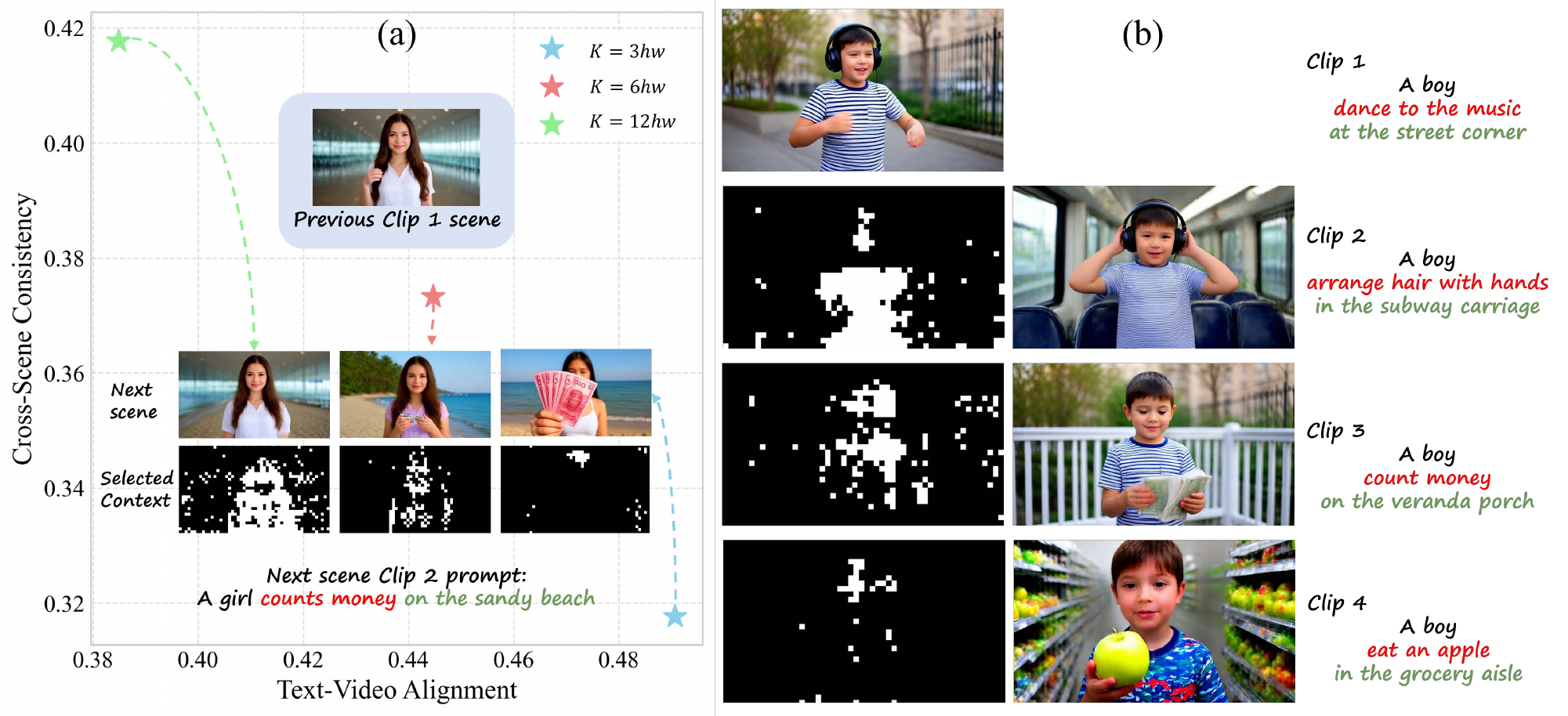}
    \vspace{-5pt}
    \caption{Analysis of context selection strategies under two settings:
(a) Varying selected context lengths with a fixed video length $L=48hw$;
(b) Fixed selection length $K=6hw$ with progressively extended video context.
}
    \label{fig:length}
\end{figure}

\textbf{The impact of context length to long video extension.}  
As shown in Figure~\ref{fig:length}(a) and Table.~\ref{tab:ab_context}, we observe that shorter context lengths prioritize tokens mainly located in facial regions—essential for preserving cross-scene identity consistency. 
This reveals that the context selection policy successfully filters irrelevant content while preserving semantically critical context, achieving optimal text-video alignment as demonstrated in Table~\ref{tab:ab_context}.
However,  insufficient selection of person-related context (below $6hw$ tokens) leads to identity drift through inconsistent facial features, as reflected by partial or inconsistent facial appearances across scenes (see Figure.~\ref{fig:length} (a)) and the low cross-scene consistency metrics presented in Table.~\ref{tab:ab_context}.
In contrast, excessive context lengths introduces distracting background elements that compromise prompt adherence. 
We show that a context length of $K=6hw$ achieves a good balance between maintaining identity consistency and adapting to evolving textual prompts.

\vspace{-5pt}
\textbf{Analysis on context selection under different video extension length.}
Given a fixed context length determined earlier, we examine how the context selection behavior evolves as the video length increases. As shown in Figure~\ref{fig:length}(b), as more clips are generated, the expanding temporal range encourages the context selection policy to focus increasingly on regions most relevant to the ongoing content generation. For instance, we observe that clips 2 to 4 progressively attend more to the body region in clip 1, which becomes semantically important for maintaining cross-scene consistency in subsequent clips.

\section{Conclusion and Limitations}
\label{sec:conclusion}
In this work, we tackle the challenge of generating coherent long-form videos guided by evolving cross-scene prompts. Existing autoregressive methods struggle with context mismanagement, often failing to preserve relevant content across scenes. To address this, we propose InfLVG, an inference-time context selection framework optimized via GRPO. It selectively retains semantically relevant context by leveraging hybrid rewards that jointly promote identity preservation, prompt alignment, and artifact suppression.
To ensure scalability, InfLVG maintains a fixed number of context tokens during generation, enabling efficient extension to longer videos without increasing computational cost. Experiments show that InfLVG achieves stronger prompt adherence and scene consistency compared to autoregressive baselines and rule-based context selection strategies.
Despite its effectiveness, our reward design remains limited in capturing broader notions of content consistency. Future work may explore leveraging vision-language models to define more generalizable and compositional consistency objectives for long video generation tasks.


\begin{thebibliography}{10}

\bibitem{Glint360k}
X.~An, X.~Zhu, Y.~Gao, Y.~Xiao, Y.~Zhao, Z.~Feng, L.~Wu, B.~Qin, M.~Zhang, D.~Zhang, et~al.
\newblock Partial fc: Training 10 million identities on a single machine.
\newblock In {\em Proceedings of the IEEE/CVF International Conference on Computer Vision}, pages 1445--1449, 2021.

\bibitem{Sketch-of-Thought}
S.~A. Aytes, J.~Baek, and S.~J. Hwang.
\newblock Sketch-of-thought: Efficient llm reasoning with adaptive cognitive-inspired sketching, 2025.

\bibitem{QWen2.5}
S.~Bai, K.~Chen, X.~Liu, J.~Wang, W.~Ge, S.~Song, K.~Dang, P.~Wang, S.~Wang, J.~Tang, et~al.
\newblock Qwen2. 5-vl technical report.
\newblock {\em arXiv preprint arXiv:2502.13923}, 2025.

\bibitem{DDPO}
K.~Black, M.~Janner, Y.~Du, I.~Kostrikov, and S.~Levine.
\newblock Training diffusion models with reinforcement learning.
\newblock {\em arXiv preprint arXiv:2305.13301}, 2023.

\bibitem{Sora}
T.~Brooks, B.~Peebles, C.~Holmes, W.~DePue, Y.~Guo, L.~Jing, D.~Schnurr, J.~Taylor, T.~Luhman, E.~Luhman, C.~Ng, R.~Wang, and A.~Ramesh.
\newblock Video generation models as world simulators.
\newblock {\em Preprint}, 2024.

\bibitem{Ouroboros}
J.~Chen, F.~Long, J.~An, Z.~Qiu, T.~Yao, J.~Luo, and T.~Mei.
\newblock Ouroboros-diffusion: Exploring consistent content generation in tuning-free long video diffusion.
\newblock {\em arXiv preprint arXiv:2501.09019}, 2025.

\bibitem{JudgeLRM}
N.~Chen, Z.~Hu, Q.~Zou, J.~Wu, Q.~Wang, B.~Hooi, and B.~He.
\newblock Judgelrm: Large reasoning models as a judge, 2025.

\bibitem{DeepSeek-r1}
DeepSeek-AI.
\newblock Deepseek-r1: Incentivizing reasoning capability in llms via reinforcement learning, 2025.

\bibitem{NOVA}
H.~Deng, T.~Pan, H.~Diao, Z.~Luo, Y.~Cui, H.~Lu, S.~Shan, Y.~Qi, and X.~Wang.
\newblock Autoregressive video generation without vector quantization.
\newblock In {\em The Thirteenth International Conference on Learning Representations}, 2025.

\bibitem{ArcFace}
J.~Deng, J.~Guo, X.~Niannan, and S.~Zafeiriou.
\newblock Arcface: Additive angular margin loss for deep face recognition.
\newblock In {\em CVPR}, 2019.

\bibitem{SD3}
P.~Esser, S.~Kulal, A.~Blattmann, R.~Entezari, J.~M{\"u}ller, H.~Saini, Y.~Levi, D.~Lorenz, A.~Sauer, F.~Boesel, et~al.
\newblock Scaling rectified flow transformers for high-resolution image synthesis.
\newblock In {\em Forty-first International Conference on Machine Learning}, 2024.

\bibitem{DPOK}
Y.~Fan, O.~Watkins, Y.~Du, H.~Liu, M.~Ryu, C.~Boutilier, P.~Abbeel, M.~Ghavamzadeh, K.~Lee, and K.~Lee.
\newblock Dpok: Reinforcement learning for fine-tuning text-to-image diffusion models.
\newblock {\em Advances in Neural Information Processing Systems}, 36:79858--79885, 2023.

\bibitem{DiT-MoE}
Z.~Fei, M.~Fan, C.~Yu, D.~Li, and J.~Huang.
\newblock Scaling diffusion transformers to 16 billion parameters.
\newblock {\em arXiv preprint arXiv:2407.11633}, 2024.

\bibitem{VideoScore}
X.~He, D.~Jiang, G.~Zhang, M.~Ku, A.~Soni, S.~Siu, H.~Chen, A.~Chandra, Z.~Jiang, A.~Arulraj, et~al.
\newblock Videoscore: Building automatic metrics to simulate fine-grained human feedback for video generation.
\newblock {\em arXiv preprint arXiv:2406.15252}, 2024.

\bibitem{DDPMs}
J.~Ho, A.~Jain, and P.~Abbeel.
\newblock Denoising diffusion probabilistic models.
\newblock {\em Advances in neural information processing systems}, 33:6840--6851, 2020.

\bibitem{visionr1}
W.~Huang, B.~Jia, Z.~Zhai, S.~Cao, Z.~Ye, F.~Zhao, Z.~Xu, Y.~Hu, and S.~Lin.
\newblock Vision-r1: Incentivizing reasoning capability in multimodal large language models.
\newblock {\em arXiv preprint arXiv:2503.06749}, 2025.

\bibitem{FIFO}
J.~Kim, J.~Kang, J.~Choi, and B.~Han.
\newblock Fifo-diffusion: Generating infinite videos from text without training.
\newblock {\em arXiv preprint arXiv:2405.11473}, 2024.

\bibitem{pickscore}
Y.~Kirstain, A.~Polyak, U.~Singer, S.~Matiana, J.~Penna, and O.~Levy.
\newblock Pick-a-pic: An open dataset of user preferences for text-to-image generation.
\newblock {\em Advances in Neural Information Processing Systems}, 36:36652--36663, 2023.

\bibitem{opensoraplan}
P.-Y. Lab and T.~A. etc.
\newblock Open-sora-plan, Apr. 2024.

\bibitem{PyFlow}
J.~Lei, X.~Hu, Y.~Wang, and D.~Liu.
\newblock Pyramidflow: High-resolution defect contrastive localization using pyramid normalizing flow.
\newblock In {\em Proceedings of the IEEE/CVF conference on computer vision and pattern recognition}, pages 14143--14152, 2023.

\bibitem{CLIPFlan}
Z.~Lin, D.~Pathak, B.~Li, J.~Li, X.~Xia, G.~Neubig, P.~Zhang, and D.~Ramanan.
\newblock Evaluating text-to-visual generation with image-to-text generation.
\newblock In {\em European Conference on Computer Vision}, pages 366--384. Springer, 2024.

\bibitem{FlowMatching}
Y.~Lipman, R.~T. Chen, H.~Ben-Hamu, M.~Nickel, and M.~Le.
\newblock Flow matching for generative modeling.
\newblock {\em arXiv preprint arXiv:2210.02747}, 2022.

\bibitem{VideoReward}
J.~Liu, G.~Liu, J.~Liang, Z.~Yuan, X.~Liu, M.~Zheng, X.~Wu, Q.~Wang, W.~Qin, M.~Xia, et~al.
\newblock Improving video generation with human feedback.
\newblock {\em arXiv preprint arXiv:2501.13918}, 2025.

\bibitem{GRADEO}
Z.~Mou, B.~Xia, Z.~Huang, W.~Yang, and J.~Jia.
\newblock Gradeo: Towards human-like evaluation for text-to-video generation via multi-step reasoning.
\newblock {\em arXiv preprint arXiv:2503.02341}, 2025.

\bibitem{s1}
N.~Muennighoff, Z.~Yang, W.~Shi, X.~L. Li, L.~Fei-Fei, H.~Hajishirzi, L.~Zettlemoyer, P.~Liang, E.~Candès, and T.~Hashimoto.
\newblock s1: Simple test-time scaling, 2025.

\bibitem{Arc2Face}
F.~Paraperas~Papantoniou, A.~Lattas, S.~Moschoglou, J.~Deng, B.~Kainz, and S.~Zafeiriou.
\newblock Arc2face: A foundation model for id-consistent human faces.
\newblock In {\em Proceedings of the European Conference on Computer Vision (ECCV)}, 2024.

\bibitem{DiT}
W.~Peebles and S.~Xie.
\newblock Scalable diffusion models with transformers.
\newblock In {\em Proceedings of the IEEE/CVF International Conference on Computer Vision}, pages 4195--4205, 2023.

\bibitem{FreeNoise}
H.~Qiu, M.~Xia, Y.~Zhang, Y.~He, X.~Wang, Y.~Shan, and Z.~Liu.
\newblock Freenoise: Tuning-free longer video diffusion via noise rescheduling.
\newblock {\em arXiv preprint arXiv:2310.15169}, 2023.

\bibitem{DPO}
R.~Rafailov, A.~Sharma, E.~Mitchell, C.~D. Manning, S.~Ermon, and C.~Finn.
\newblock Direct preference optimization: Your language model is secretly a reward model.
\newblock {\em Advances in Neural Information Processing Systems}, 36:53728--53741, 2023.

\bibitem{ragain2018choosing}
S.~Ragain and J.~Ugander.
\newblock Choosing to rank.
\newblock {\em arXiv preprint arXiv:1809.05139}, 2018.

\bibitem{Laion-5B-AesScore}
C.~Schuhmann, R.~Beaumont, R.~Vencu, C.~Gordon, R.~Wightman, M.~Cherti, T.~Coombes, A.~Katta, C.~Mullis, M.~Wortsman, et~al.
\newblock Laion-5b: An open large-scale dataset for training next generation image-text models.
\newblock {\em Advances in neural information processing systems}, 35:25278--25294, 2022.

\bibitem{ppo}
J.~Schulman, F.~Wolski, P.~Dhariwal, A.~Radford, and O.~Klimov.
\newblock Proximal policy optimization algorithms.
\newblock {\em arXiv preprint arXiv:1707.06347}, 2017.

\bibitem{grpo}
Z.~Shao, P.~Wang, Q.~Zhu, R.~Xu, J.~Song, X.~Bi, H.~Zhang, M.~Zhang, Y.~Li, Y.~Wu, et~al.
\newblock Deepseekmath: Pushing the limits of mathematical reasoning in open language models.
\newblock {\em arXiv preprint arXiv:2402.03300}, 2024.

\bibitem{DDIM}
J.~Song, C.~Meng, and S.~Ermon.
\newblock Denoising diffusion implicit models.
\newblock {\em arXiv preprint arXiv:2010.02502}, 2020.

\bibitem{Mochi}
G.~Team.
\newblock Mochi 1.
\newblock \url{https://github.com/genmoai/models}, 2024.

\bibitem{Gemini}
G.~Team, R.~Anil, S.~Borgeaud, J.-B. Alayrac, J.~Yu, R.~Soricut, J.~Schalkwyk, A.~M. Dai, A.~Hauth, K.~Millican, et~al.
\newblock Gemini: a family of highly capable multimodal models.
\newblock {\em arXiv preprint arXiv:2312.11805}, 2023.

\bibitem{PlackettLuce}
H.~L. Turner, J.~{van Etten}, D.~Firth, and I.~Kosmidis.
\newblock Modelling rankings in {R}: The {PlackettLuce} package.
\newblock {\em Computational Statistics}, 35:1027--1057, 2020.

\bibitem{Diffusion-DPO}
B.~Wallace, M.~Dang, R.~Rafailov, L.~Zhou, A.~Lou, S.~Purushwalkam, S.~Ermon, C.~Xiong, S.~Joty, and N.~Naik.
\newblock Diffusion model alignment using direct preference optimization.
\newblock In {\em Proceedings of the IEEE/CVF Conference on Computer Vision and Pattern Recognition}, pages 8228--8238, 2024.

\bibitem{WanX}
T.~Wan.
\newblock Wan: Open and advanced large-scale video generative models, 2025.

\bibitem{InternVid}
Y.~Wang, Y.~He, Y.~Li, K.~Li, J.~Yu, X.~Ma, X.~Chen, Y.~Wang, P.~Luo, Z.~Liu, Y.~Wang, L.~Wang, and Y.~Qiao.
\newblock Internvid: A large-scale video-text dataset for multimodal understanding and generation.
\newblock {\em arXiv preprint arXiv:2307.06942}, 2023.

\bibitem{CoT-Survey-Hao-Fei}
Y.~Wang, S.~Wu, Y.~Zhang, S.~Yan, Z.~Liu, J.~Luo, and H.~Fei.
\newblock Multimodal chain-of-thought reasoning: A comprehensive survey, 2025.

\bibitem{HPSv2}
X.~Wu, Y.~Hao, K.~Sun, Y.~Chen, F.~Zhu, R.~Zhao, and H.~Li.
\newblock Human preference score v2: A solid benchmark for evaluating human preferences of text-to-image synthesis.
\newblock {\em arXiv preprint arXiv:2306.09341}, 2023.

\bibitem{wu2024boosting}
X.~Wu, S.~Huang, G.~Wang, J.~Xiong, and F.~Wei.
\newblock Boosting text-to-video generative model with mllms feedback.
\newblock In {\em The Thirty-eighth Annual Conference on Neural Information Processing Systems}, 2024.

\bibitem{VideoPrefer}
X.~Wu, S.~Huang, G.~Wang, J.~Xiong, and F.~Wei.
\newblock Boosting text-to-video generative model with mllms feedback.
\newblock In {\em The Thirty-eighth Annual Conference on Neural Information Processing Systems}, 2024.

\bibitem{hps}
X.~Wu, K.~Sun, F.~Zhu, R.~Zhao, and H.~Li.
\newblock Human preference score: Better aligning text-to-image models with human preference.
\newblock In {\em Proceedings of the IEEE/CVF International Conference on Computer Vision}, pages 2096--2105, 2023.

\bibitem{ImageReward}
J.~Xu, X.~Liu, Y.~Wu, Y.~Tong, Q.~Li, M.~Ding, J.~Tang, and Y.~Dong.
\newblock Imagereward: Learning and evaluating human preferences for text-to-image generation.
\newblock {\em Advances in Neural Information Processing Systems}, 36, 2024.

\bibitem{CogVideoX}
Z.~Yang, J.~Teng, W.~Zheng, M.~Ding, S.~Huang, J.~Xu, Y.~Yang, W.~Hong, X.~Zhang, G.~Feng, et~al.
\newblock Cogvideox: Text-to-video diffusion models with an expert transformer.
\newblock {\em arXiv preprint arXiv:2408.06072}, 2024.

\bibitem{CausVid}
T.~Yin, Q.~Zhang, R.~Zhang, W.~T. Freeman, F.~Durand, E.~Shechtman, and X.~Huang.
\newblock From slow bidirectional to fast causal video generators.
\newblock {\em arXiv preprint arXiv:2412.07772}, 2024.

\bibitem{InstructVideo}
H.~Yuan, S.~Zhang, X.~Wang, Y.~Wei, T.~Feng, Y.~Pan, Y.~Zhang, Z.~Liu, S.~Albanie, and D.~Ni.
\newblock Instructvideo: Instructing video diffusion models with human feedback.
\newblock In {\em Proceedings of the IEEE/CVF Conference on Computer Vision and Pattern Recognition}, pages 6463--6474, 2024.

\bibitem{BigBird}
M.~Zaheer, G.~Guruganesh, K.~A. Dubey, J.~Ainslie, C.~Alberti, S.~Ontanon, P.~Pham, A.~Ravula, Q.~Wang, L.~Yang, et~al.
\newblock Big bird: Transformers for longer sequences.
\newblock {\em Advances in neural information processing systems}, 33:17283--17297, 2020.

\bibitem{WebFace42M}
Z.~Zhu, G.~Huang, J.~Deng, Y.~Ye, J.~Huang, X.~Chen, J.~Zhu, T.~Yang, J.~Lu, D.~Du, and J.~Zhou.
\newblock Webface260m: A benchmark unveiling the power of million-scale deep face recognition.
\newblock In {\em Proceedings of the IEEE/CVF Conference on Computer Vision and Pattern Recognition (CVPR)}, pages 10492--10502, June 2021.

\end{thebibliography}

\newpage

\appendix
\section{Details of CsVBench}
The generation of consistent cross-scene long videos can be approached in two ways: (1) naively extending a single scene to produce longer, temporally stable content, or (2) composing multiple scenes featuring the same subject for narrative progression. 
However, single-scene extensions often lack semantic richness. In this work, we focus on generating coherent long videos across multiple scenes. Nonetheless, our method remains effective for single-scene scenarios, as shown in Figure~\ref{fig:demo_single_multi}.

To evaluate cross-scene video generation in terms of visual quality, cross-scene consistency, and text–video alignment, we propose a new benchmark, Cross-scene Video Benchmark (CsVBench). This benchmark includes a diverse set of prompts designed for evaluation, along with quantitative metrics for systematic analysis. Details are provided below.

\subsection{Evaluation Prompts Curation}
CsVBench evaluates models using prompts that describe cross-scene events, where a consistent subject performs different actions across varied backgrounds.
Each unique combination of \textbf{H:[human]}, \textbf{A:[action]}, and \textbf{B:[background]} is defined as an Event Prompt, and prompts sharing the same subject are grouped together into an \textbf{Event Prompt Set (EPS)}.

For each EPS, we fix a single subject identity and randomly sample $S = 4$ distinct action-background pairs to construct a set of event prompts. The EPS for $i$-th identity is formally defined as
$$
    \text{EPS}_i = \big\{ \big( H_i, A_{j_s}, B_{k_s} \big) \mid s \in \{1,\dots,S\}, \, j_s, k_s \sim \mathcal{U}\{0, S-1\} \big\}.
$$

We curate a total of $1000$ EPSs which are manually filtered to ensure semantic validity and diversity, resulting in the final CsVBench benchmark.

Examples of EPSs used in our experiments are presented below:

\begin{lstlisting}[
  caption={16 Examples of EPSs used in our experiments.},
  label={lst:eps_examples},
  language=python,
  basicstyle=\ttfamily\small,
  breaklines=true,
  backgroundcolor=\color{gray!10},
  numbers=left,
  xleftmargin=1em,
  xrightmargin=0em
]

{
  "0": "A man play with a dog at the train platform",
  "1": "A man play a guitar at the bus stop",
  "2": "A man write in a notebook at the coffee shop",
  "3": "A man carry a basket of fruit in the school hallway",
  "4": "A man drinking water in the kitchen corner",
  "5": "A man wipe away tears by the riverbank path"
}
{
  "0": "An elder write in a notebook at the bus stop",
  "1": "An elder carry a basket of fruit at the coffee shop",
  "2": "An elder drinking water in the school hallway",
  "3": "An elder wipe away tears in the kitchen corner",
  "4": "An elder make a telephone call by the riverbank path",
  "5": "An elder eat an apple in the park pavilion"
}
{
  "0": "An elder make a telephone call in the school hallway",
  "1": "An elder eat an apple in the kitchen corner",
  "2": "An elder dance to the music by the riverbank path",
  "3": "An elder arrange hair with hands in the park pavilion",
  "4": "An elder count money in the grocery aisle",
  "5": "An elder scratch head at the gas station"
}

{
  "0": "A boy dance to the music in the kitchen corner",
  "1": "A boy arrange hair with hands by the riverbank path",
  "2": "A boy count money in the park pavilion",
  "3": "A boy scratch head in the grocery aisle",
  "4": "A boy sitting down at the gas station",
  "5": "A boy standing up at the hotel lobby"
}
{
  "0": "A policeman holding an umbrella in the grocery aisle",
  "1": "A policeman play with a cat at the gas station",
  "2": "A policeman play with a dog at the hotel lobby",
  "3": "A policeman play a guitar in the hotel room",
  "4": "A policeman write in a notebook on the bike lane",
  "5": "A policeman carry a basket of fruit in the art gallery"
}
{
  "0": "A nurse drinking water in the hotel room",
  "1": "A nurse wipe away tears on the bike lane",
  "2": "A nurse make a telephone call in the art gallery",
  "3": "A nurse eat an apple at the hospital ward",
  "4": "A nurse dance to the music on the fishing pier",
  "5": "A nurse arrange hair with hands at the poolside lounge"
}
{
  "0": "Elder wearing a wide-brimmed hat eat an apple at the city fountain",
  "1": "Elder wearing a wide-brimmed hat dance to the music on the church steps",
  "2": "Elder wearing a wide-brimmed hat arrange hair with hands by the garden gate",
  "3": "Elder wearing a wide-brimmed hat count money in the coffee queue",
  "4": "Elder wearing a wide-brimmed hat scratch head on the mountain summit",
  "5": "Elder wearing a wide-brimmed hat sitting down at the convention center"
}
{
  "0": "Elder wearing a wide-brimmed hat scratch head by the garden gate",
  "1": "Elder wearing a wide-brimmed hat sitting down in the coffee queue",
  "2": "Elder wearing a wide-brimmed hat standing up on the mountain summit",
  "3": "Elder wearing a wide-brimmed hat holding an umbrella at the convention center",
  "4": "Elder wearing a wide-brimmed hat play with a cat in the concert hall",
  "5": "Elder wearing a wide-brimmed hat play with a dog on the running track"
}
{
  "0": "A woman standing up in the coffee queue",
  "1": "A woman holding an umbrella on the mountain summit",
  "2": "A woman play with a cat at the convention center",
  "3": "A woman play with a dog in the concert hall",
  "4": "A woman play a guitar on the running track",
  "5": "A woman write in a notebook at the subway entrance"
}
{
  "0": "An elder write in a notebook on the tennis court",
  "1": "An elder carry a basket of fruit at the soccer stadium",
  "2": "An elder drinking water in the waiting room",
  "3": "An elder wipe away tears on the hiking path",
  "4": "An elder make a telephone call at the night market",
  "5": "An elder eat an apple in the music studio"
}
{
  "0": "A boy drinking water at the soccer stadium",
  "1": "A boy wipe away tears in the waiting room",
  "2": "A boy make a telephone call on the hiking path",
  "3": "A boy eat an apple at the night market",
  "4": "A boy dance to the music in the music studio",
  "5": "A boy arrange hair with hands on the ferry deck"
}
{
  "0": "A nurse play with a dog at the bakery counter",
  "1": "A nurse play a guitar in the dance studio",
  "2": "A nurse write in a notebook on the golf course",
  "3": "A nurse carry a basket of fruit on the airport tarmac",
  "4": "A nurse drinking water on the bookstore floor",
  "5": "A nurse wipe away tears in the hospital waiting area"
}
{
  "0": "A nurse sitting down in the music studio",
  "1": "A nurse standing up on the ferry deck",
  "2": "A nurse holding an umbrella at the bakery counter",
  "3": "A nurse play with a cat in the dance studio",
  "4": "A nurse play with a dog on the golf course",
  "5": "A nurse play a guitar on the airport tarmac"
}
{
  "0": "A girl wipe away tears on the running track",
  "1": "A girl make a telephone call at the subway entrance",
  "2": "A girl eat an apple in the flower market",
  "3": "A girl dance to the music on the skateboard ramp",
  "4": "A girl arrange hair with hands at the farmer's market",
  "5": "A girl count money in the wine cellar"
}
{
  "0": "A man drinking water on the golf course",
  "1": "A man wipe away tears on the airport tarmac",
  "2": "A man make a telephone call on the bookstore floor",
  "3": "A man eat an apple in the hospital waiting area",
  "4": "A man dance to the music on the bridge railing",
  "5": "A man arrange hair with hands at the zoo entrance"
}
{
  "0": "A policeman count money by the riverbank path",
  "1": "A policeman scratch head in the park pavilion",
  "2": "A policeman sitting down in the grocery aisle",
  "3": "A policeman standing up at the gas station",
  "4": "A policeman holding an umbrella at the hotel lobby",
  "5": "A policeman play with a cat in the hotel room"
}
\end{lstlisting}


\subsection{ Metrics Details}
To quantitatively evaluate the visual quality and the adherence to both text and visual cues in our method, we employ a range of metrics on the generated videos:
\begin{itemize}
\item \textbf{Video Quality}: We evaluate video quality using HPSv2~\cite{HPSv2} and an Aesthetic Score~\cite{Laion-5B-AesScore} to measure the human preference and aesthetic quality of video frames. 
To compute the overall video score, we evaluated these metrics on uniformly sampled key frames (extracted at 4-frame intervals) and subsequently averaged the resulting scores.
\item \textbf{Text-Video Alignment}: We employ CLIP-Flan-T5-xl~\cite{CLIPFlan} and ViCLIP~\cite{InternVid} to quantify text-video alignment through cosine similarity measurement, as this semantic correspondence represents a critical dimension of video generation quality. The evaluation is particularly relevant for autoregressive models, where maintaining consistent alignment throughout long sequences presents unique technical challenges.
\item \textbf{Cross-scene Consistency}: We also assess the consistency of human identity across different scenes, with a specific focus on face consistency. For this purpose, we employ two versions of ArcFace: ArcFace-360k~\cite{ArcFace}, trained on Glint360k~\cite{Glint360k}, and ArcFace-42M~\cite{Arc2Face}, trained on WebFace42M~\cite{WebFace42M}.
\end{itemize}
Furthermore,  QWen2.5-VL~\cite{QWen2.5} is utilized to evaluate the aforementioned aspects due to its promising multimodal comprehension capabilities, which provides a score to judge the presence of color mosaic artifacts in the generated videos, the degree of text-video alignment, and the consistency of human identity in cross-scene videos with specified instructions.

\begin{figure}[htpb]
    \centering
    \includegraphics[width=0.4\linewidth]{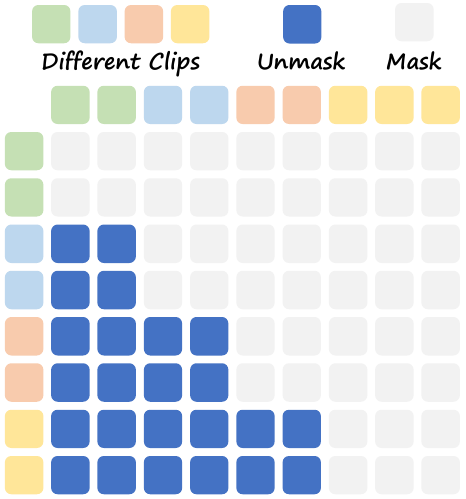}
    \caption{Illustration of clip-wise cross-scene cosine similarity calculation.}
    \label{fig:supp_mask}
\end{figure}

To evaluate cross-scene consistency, we compute pairwise similarities between frame-level features extracted by arcFace across different video clips. Since the videos are autoregressively generated, we specifically focus on comparing each clip with its preceding counterparts while omitting self-similarity (i.e., the similarity of a frame with itself). 
Formally, the cross-scene consistency SIM is calculated by comparison of frames $X$ at different clips, which is formulated as:
\begin{equation}
    \label{eq:cos_sim}
    \text{SIM} = \text{AVG}({\hat{X}\hat{X}^T} \odot M), \quad \hat{X} = \frac{X}{|X|} \in \mathbb{R}^{F \times C},\; M \in \mathbb{R}^{F \times F}
\end{equation}
where $F$ denotes the number of video frames from all video clips and $C$ is the embedding dimension of each frame.  
$M$ is a specially designed binary mask (visualized in Figure~\ref{fig:supp_mask}) that eliminates invalid similarity computations, including self-comparisons and non-causal temporal relationships.
$\text{AVG}(\cdot)$ computes the mean over all valid (unmasked) similarity scores.

\section{Experiment Setup Details}

\subsection{Artifact Reward}

As shown in ~\ref{fig:ab_reward}, artifact reward $r_\text{artifact}$ is calculated by  QWen2.5-VL to act as an additional verifier for video artifacts detection.
This VLM is prompted with a carefully designed system and user query to determine whether such artifacts are present in the generated frames. The resulting judgment is incorporated as a reward to guide the optimization of context selection .

\begin{figure}[htpb]
    \centering
    \includegraphics[width=0.4\linewidth]{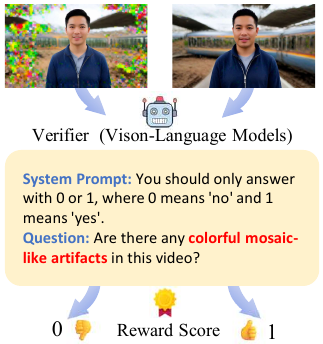} 
    \caption{Visualization of mosaic color artifact detection via VLM~\cite{QWen2.5} and artifact reward $r_\text{artifact}$ computation.}
    \label{fig:mosaic}
\end{figure}

\subsection{Vanilla Extension Implementation}
\begin{figure}[htpb]
    \centering
    \includegraphics[width=\linewidth]{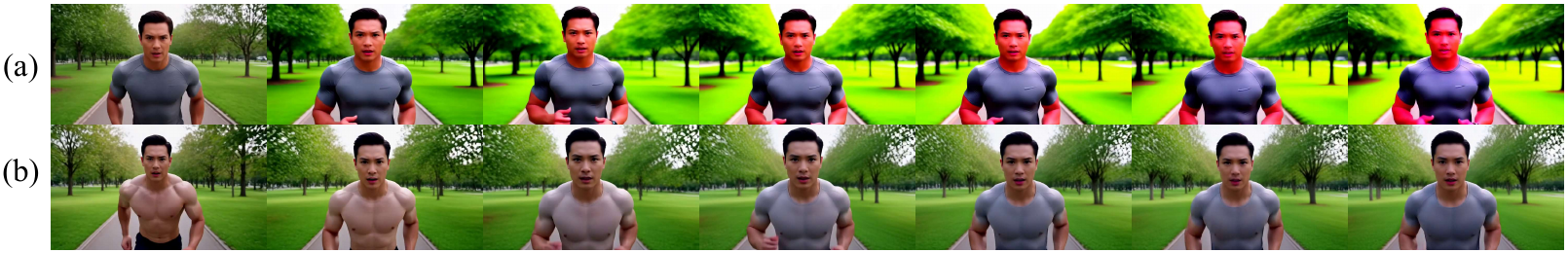}
    \caption{
      Vanilla autoregressive long video extension with different implementations .
      (a) Official rollout implementation results in progressive quality degradation, especially noticeable as increasing saturation of facial colors.
      (b) Our vanilla extension implementation denoises current frame by the KV cache at corresponding denoising timestep to generate the next frame, resulting in almost no degradation—only minor clothing variations, which stablizes the base model’s performance in vanilla extension model variant in our experiment comparison.
    }
    \label{fig:supp_rollout}
\end{figure}

In our experiment, we compare with the vanilla extension method that conduct video extension conditioning on kv cache generated from all preceding frames.
Notably, we do not utilize the official rollout implementation in the CausVid repository for generating long videos, as this approach can lead to significant quality degradation. In the vanilla extension process, each new frame can attend only to the last denoising timestep of the previous frame’s KV cache, which leads to severe degration of visual quality  as illustrated in Figure \ref{fig:supp_rollout} (a).
We thus turn to denoise each noisy frame with kv cache aggregated from preceding frames of corresponding denoising timesteps and alleviate the quality degradation significantly.

\section{Video Extension on Generalized Subjects}
\begin{figure}[htpb]
    \centering
    \includegraphics[width=\linewidth]{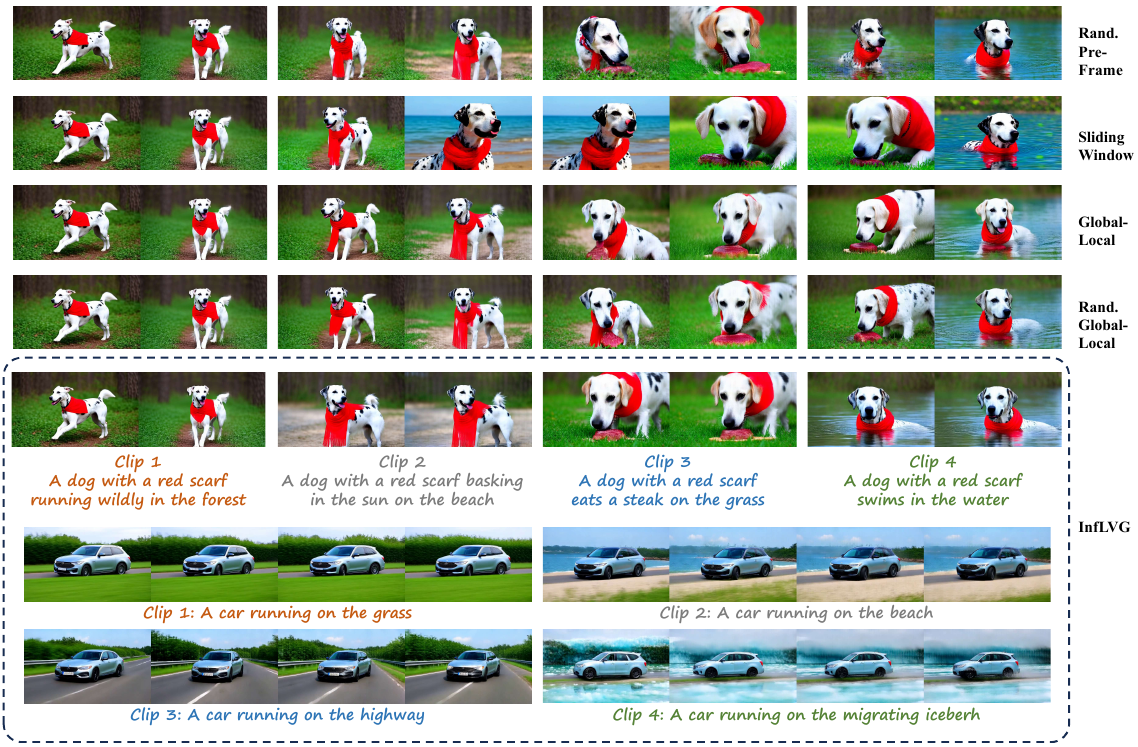}
    \caption{InfLVG generates consistent long videos of generalized subjects (e.g., dogs, cars) across scenes through our DINOv2-based cross-scene consistency reward.}
    \label{fig:supp_animal}
\end{figure}

The proposed InfLVG incorporates hybrid rewards to jointly optimize three key aspects: video quality, cross-scene consistency, and text-video alignment. While our primary experiments focus on human-centric scenarios using ArcFace for identity consistency evaluation, the framework is fundamentally domain-agnostic.
InfLVG demonstrates strong generalization capability across diverse subject domains, including animals (e.g., dogs) and objects (e.g., cars). This flexibility is achieved through our adaptive content consistency measurement approach: (1) using SAM for precise subject region detection in generated videos, followed by (2) DINOv2-based feature embedding extraction to compute the content consistency reward $r_\text{content}$ through cosine similarity.
As evidenced in Figure~\ref{fig:supp_animal}, InfLVG outperforms comparative methods in maintaining both scene-to-scene consistency (for the dog subject) and text-to-video alignment fidelity. The framework's effectiveness extends beyond animate subjects, showing equally robust performance on object categories like cars.

\section{Safeguards and Social Impact}

We manually filtered Not-Safe-For-Work (NSFW) content in our CsVBench. Therefore, users employing our standard CsVBench should not encounter the generation of NSFW content by video generation models. On the positive side, our approach has the potential to advance video generation technologies for various applications, including long video extension and storytelling. However, we also recognize the potential for misuse, such as the generation of deepfakes for disinformation or surveillance.

\end{document}